\documentclass[runningheads]{llncs}
\usepackage{graphicx}

\usepackage{amsthm}
\usepackage{amsmath}
\usepackage{amsfonts}
\usepackage{bbm}

\usepackage{graphicx}
\usepackage{subcaption}

\usepackage{todonotes}
\usepackage[misc]{ifsym}

\usepackage{geometry}
\usepackage{hyperref}
\usepackage{url}

\usepackage[square,numbers]{natbib}
\PassOptionsToPackage{numbers, compress}{natbib}
\begin{document}
\title{Training Data Attribution for Diffusion Models}%
\author{Zheng Dai\inst{1}(\Letter)\orcidID{0000-0002-8828-1075} \and
David K Gifford\inst{1}(\Letter)\orcidID{0000-0003-1709-4034}}
\authorrunning{Z. Dai et al.}%
\institute{Computer Science and Artificial Intelligence Laboratory, Massachusetts Institute of Technology, Cambridge MA 02139, USA \email{\{zhengdai,gifford\}@mit.edu}}
\maketitle              %
\begin{abstract}
Diffusion models have become increasingly popular for synthesizing high-quality samples based on training datasets. However, given the oftentimes enormous sizes of the training datasets, it is difficult to assess how training data impact the samples produced by a trained diffusion model.  The difficulty of relating diffusion model inputs and outputs poses significant challenges to model explainability and training data attribution. Here we propose a novel solution that reveals how training data influence the output of diffusion models through the use of ensembles.  In our approach individual models in an \emph{encoded ensemble} are trained on carefully engineered splits of the overall training data to permit the identification of influential training examples.  The resulting model ensembles enable efficient ablation of training data influence, allowing us to assess the impact of training data on model outputs. We demonstrate the viability of these ensembles as generative models and the validity of our approach to assessing influence.

\end{abstract}
\newcommand{\prob}[1]{\textrm{Pr}(#1)}
\newcommand{\linf}[1]{\| #1 \|_{\infty}}

\newcommand{\codingensemble}{encoded ensemble}
\newcommand{\Codingensemble}{Encoded ensemble}
\newcommand{\acodingensemble}{an encoded ensemble}
\newcommand{\Acodingensemble}{An encoded ensemble}
\section{Introduction}

Diffusion models have emerged as powerful tools for modeling and sampling from complex natural distributions. These models, fueled by their remarkable performance, have garnered significant attention and achieved remarkable results in a wide array of applications ranging from text conditioned image generation \cite{ramesh2022hierarchical}, video generation \cite{ho2022video}, audio synthesis \cite{zhang2023survey}, and even therapeutic design \cite{luo2022antigen}. To attain these performances, these models often need to be trained on massive corpuses of training data, which makes it challenging to assess the influence of the training samples. Assessing the value of training samples is highly valuable due to its many applications in fields including model interpretability \cite{koh2017understanding}, machine unlearning \cite{nguyen2022survey}, data poisoning \cite{chen2017targeted}, fairness \cite{mehrabi2021survey}, and privacy \cite{shokri2017membership}.

We present \emph{\codingensemble{}s}, a method for training data attribution through the use of ensembles of diffusion models trained on specially coded splits of the training data. To assess the influence of a training point on a generated sample, we simply remove models that have seen the training point from the ensemble and regenerate the sample with the ablated ensemble. We empirically demonstrate the effectiveness of this method. Furthermore, we derive an approximate method that forgoes the need to regenerate a sample, enabling the search for influential training data across entire training sets, which would otherwise be intractable.

\subsection{Related work}

Diffusion models were originally introduced to machine learning by \citet{sohl2015deep}, and were improved by \citet{ho2020denoising}, whose models we base our models on.

The attribution of model outputs to training data can be approximated using influence functions \cite{koh2017understanding}. Calculating training data influence is an active area of research within the machine unlearning community \cite{nguyen2022survey}. Influence functions approximate the effect of removing a data point from the training set by reducing its weight in the training set by an infinitesimal amount, and approximating the resulting parameters via Taylor expansion. While informative, the true meaning of these approximations are sometimes disputed \cite{basu2020influence}. Therefore, we opt against this approach, especially in the
absence of prior work that establishes a good ground truth to benchmark against in the case of generative diffusion models.

The use of ensembling to boost model predictions is a common technique in machine learning \cite{sagi2018ensemble}, though uncommon in generative models. The only prior work involving both ensembles and diffusion models to the authors knowledge comes from \citet{balaji2022ediffi}, which differs significantly from our work.  Firstly, their aim is to increase model performance, rather than enabling data attribution. Secondly, they run their ensembles in sequence, so each denoising step is still taken by a single model, while ours ensembles the outputs for each denoising step.

\subsection{Our Contributions}

Our contributions are the following:
\begin{enumerate}
    \item To the author's knowledge, this is the first work quantifying training data influence in diffusion models.
    \item We present an effective strategy for temporary unlearning to assess influence by ensemble ablation. Our method leverages the fact that the unlearning is only needed temporarily, and is therefore much more efficient than traditional unlearning methods.
    \item We derive an approximation scheme that allows us to efficiently assess influence over entire training sets.
    \item We demonstrate empirically that an ensemble of diffusion models is a viable generative model, and that ensemble ablation does produce a viable measure of influence.
\end{enumerate}
\section{Methods}

\subsection{Preliminaries}

This subsection details all the notation used in this work. The reader is recommended to skip this on first reading and refer back whenever unknown notation is encountered. Proofs to any theorems are supplied in Appendix \ref{appdx-proofs}.

Given a vector (or higher dimensional tensor) $v$, the \emph{Hamming weight} of the vector $v$ is the number of non-zero entries of that vector. For bit-vectors, this is effectively the number of 1s. We will also use $\|v\|_{\infty}$ to denote the max-norm of $v$, which is the magnitude of the entry in $v$ most distant from 0. Vectors will be 1-indexed unless otherwise stated.

Given a set $X$, we will use $2^{X}$ to denote its power set, the set of all subsets of $X$. We will use $\emptyset$ to denote the empty set. If $D$ is some probability distribution over $X$, then we will use $x \sim D$ to denote ``$x$ is drawn from $D$''.

\subsection{We assess training data influence by generating counterfactuals through temporary unlearning}

The key question we would like to answer is: \emph{``what is the influence of a piece of training data over a given generated sample?''}
We operationalize this question by considering the counterfactual: \emph{``if the model had not been trained on this piece of training data, how different would the model output look?''}
This is the basis of \emph{leave-one-out retraining}, where the idea is to retrain a model with a dataset that has the training sample in question removed, and then observing how the output of the retrained model differs from the original one given the same input. In practise, unlearning approaches are often used to simulate the effects of retraining \cite{koh2017understanding, hammoudeh2022training}.
We apply this approach to generative models. Although in abstract generative models only define a distribution over the output space and therefore lack an input domain, in practice they are implemented as functions that map random noise to samples. We will refer to the noise used as input to generate a particular sample as the \emph{exogenous noise}.

Our strategy is therefore the following: given a generated sample, the exogenous noise used to generate it, and the training sample we wish to assess for influence, we first unlearn the piece of training data from the model. We then input the exogenous noise into the new model, generating a \emph{counterfactual sample}. The original and counterfactual samples can then be compared with a variety of methods.

The main challenge arises from the need to unlearn training data. The most foolproof method of unlearning data is of course to retrain from a model from scratch without including the unlearned data. This is expensive, especially for diffusion models, which we target \cite{ho2020denoising, ramesh2022hierarchical}. One possible solution is to turn to approximate methods such as \cite{koh2017understanding}. However we opt against these approximate methods because it is difficult to assess their accuracy when applied in the context of diffusion model counterfactuals, especially in the absence of prior work that establishes a good ground truth to benchmark against. As we will see, the influence of training data can be unintuitive\footnote{For example, see Figure \ref{fig-databaseSearch}.}.

\begin{figure}

\begin{center}
    \includegraphics[width=0.8\textwidth]{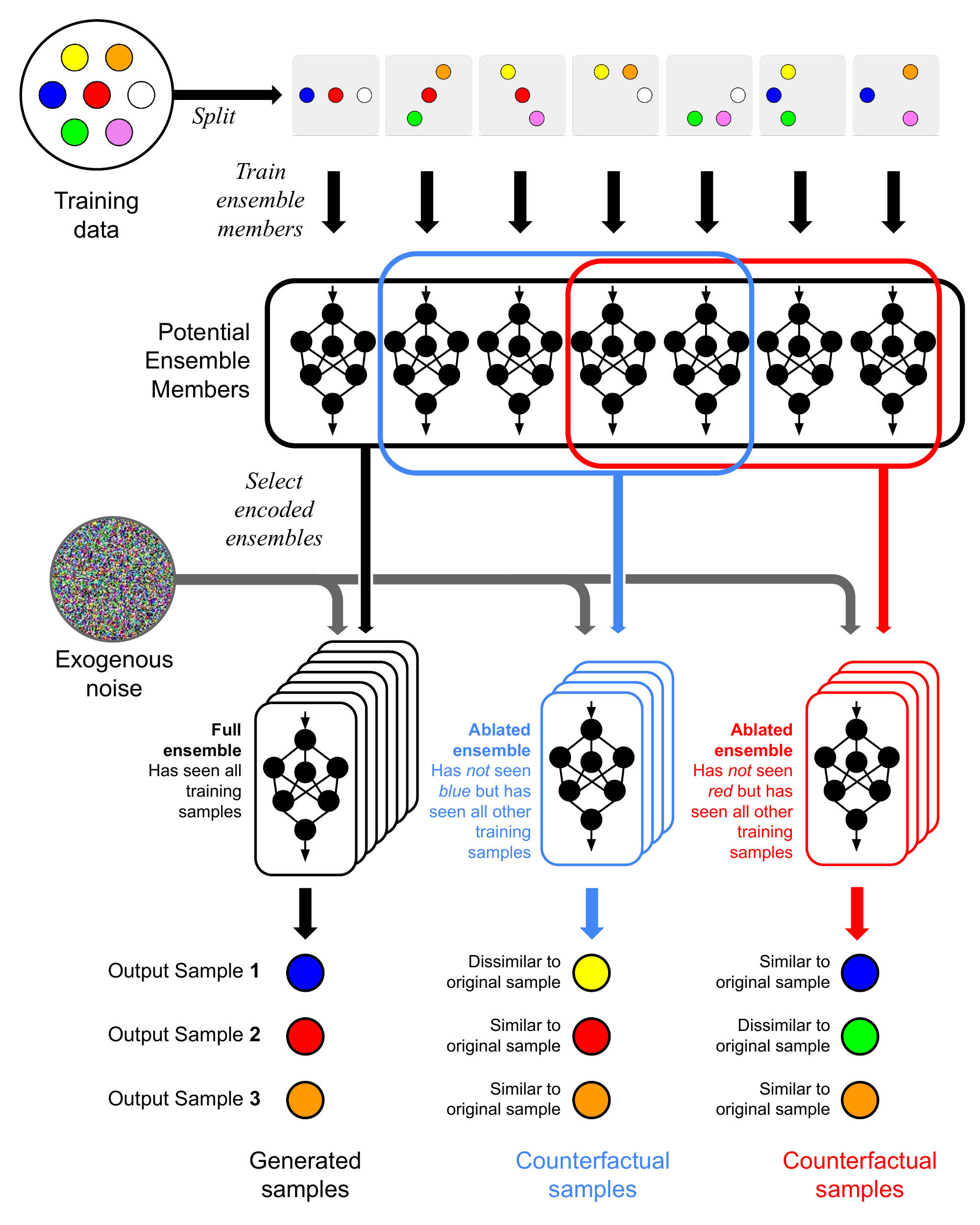}
\end{center}
 
  \caption{Inducing temporary unlearning in \codingensemble{}s of generative models to generate counterfactuals}
  \label{fig-overview}
  We train a \codingensemble{} of generative models on carefully engineered splits of the traing data.
  We split our data such that, for each piece of training data, there exists some subset of the trained models in the ensemble that has, collectively, seen every other piece of training data except the one in question. This allows us to efficiently induce unlearning by ablating the ensemble. These ablated ensembles can then be used to generate counterfactual samples, which can then be compared to the original samples downstream to assess influence.
\end{figure}

Our solution is to train an ensemble of models on varying subsets of the training data. To unlearn a specific sample from the training set, it suffices to remove all the models that have seen the sample from the ensemble (see Figure \ref{fig-overview}). Since we are only concerned with assessing influence, we need only induce unlearning \emph{temporarily}, allowing us to circumvent the main challenges associated with unlearning. Therefore, there is no need to retrain the removed models, making temporary unlearning an instantaneous operation.

\subsection{\Codingensemble{}s allow us to induce temporary unlearning}
\label{sec-method-main}

We will think of a class of diffusion models as a function $f$ that takes three inputs: $x$, $t$, and $\theta$. $x$ is the input, $t$ is the time along the diffusion schedule, and $\theta$ represents the trainable parameters.

\newcommand{\trainset}{X}
\newcommand{\traindistrs}{\mathcal{X}}
\newcommand{\remvpt}{\tilde{x}}

Let $\trainset$ denote the total training data available, and let $\traindistrs$ denote the uniform distribution over $2^{\trainset}$. Let $\mathcal{A}$ denote a training procedure that takes in two inputs, where the first is a set of training samples and the second is exogenous noise\footnote{We have previously defined exogenous noise as input to a generative model. We can think of a training algorithm, once curried with training data, as a generative model that defines a distribution over the model parameter space. Hence the use of the term ``exogenous noise'' is identical in this context.}. The training procedure then outputs a $\theta$ that can be used as the third argument to $f$.

Our proposal is to use an ensemble of diffusion models as our denoiser. We will denote this denoiser as $f_e$, which we formally define as:

\begin{equation}
\label{eqn-ens}
    f_e(x,t) = \mathbb{E}_{S \sim \traindistrs}[
    \mathbb{E}_{r \sim R} [
    f(x,t,
    \mathcal{A}(S, r)
    )
    ]]
\end{equation}

Where $R$ is some distribution over the exogenous noise that is used in the training process. Given this, if we consider the counterfactual dataset where some point $\remvpt$ did not exist in $\trainset$, the denoiser would instead be:

\begin{equation}
\label{eqn-abl}
    f_e^{-\remvpt}(x, t) = \frac{1}{\prob{x \in S' \sim \traindistrs}}\mathbb{E}_{S \sim \traindistrs}[
    \mathbb{E}_{r \sim R} [
    f(x,t,
    \mathcal{A}(S, r)
    )\mathbbm{1}_{\remvpt \notin S}
    ]]
\end{equation}

So all that needs to be done is to remove the models that have seen $\tilde{x}$ and normalize the output. To evaluate the denoiser exactly would require training over all subsets of the training data which is intractable. Therefore, our approach is to carefully engineer splits of the training data, such that the splits encode the training data in a way that allows us to surgically remove the influence of any one training sample by removing models.

We use the following strategy: first, we assign, uniformly at random, to each sample in the training set a unique code given by a bit vector of prespecified length $n$ with some prespecified hamming weight $h$. We then create $n$ training subsets, such that the $i$th training set contains exactly the samples whose bit vector is 1 at position $i$. These $n$ training sets are then used to train $n$ models, which then form the members of the ensemble. The outputs of the models are averaged to produce the outputs of the ensemble. Since subsets of this ensemble encode training data, we will refer to this as a \emph{\codingensemble{}}.

Given a training dataset of size $|X|$, we can assign these bit vectors as long as $\binom{n}{h} \geq |X|$, so only $\mathcal{O}(log(|X|)^{1 + \epsilon})$ models need to be trained\footnote{for arbitrarily small non-zero $\epsilon$}. This assignment provides several desirable guarantees:

\begin{enumerate}
    \item For each sample in the training data, there exists at least one subset of the models of the \codingensemble{} that has not seen that sample, but has seen every other sample.
    \item Given some conditions, the \codingensemble{} is an asymptotically unbiased estimator of $f_e$ as $n$ grows larger. Similarly, when models that have seen some sample $\remvpt$ is removed from the \codingensemble{}, it becomes an unbiased estimator of $f_e^{-\remvpt}$.
\end{enumerate}

More formally, we can state the above as the following:

\begin{theorem}
\label{thm-main}
    If $n$ models are trained on $n$ subsets of the dataset $\trainset$ according to the assignment described to produce \acodingensemble{}, the following holds:
    \begin{enumerate}
        \item For any $x$ in the training set, if we remove all models that have been trained on $x$, for all other $x'$ there exists at least one model that has not been removed that has been trained on $x'$.
    \end{enumerate}
    Furthermore, if the bit vectors are assigned uniformly at random, and additionally $\binom{n}{h} \geq 2|\trainset|$, $n = 2h$, and $\linf{f(x',t',\theta')} \leq C$ for all inputs $x'$, $t'$, and $\theta'$ we have for arbitrary $\remvpt \in \trainset$:
    \begin{enumerate}
        \item $\|\mathbb{E}[\hat{f_e}(x,t) - f_e(x,t)]\|_{\infty} \leq ln(16)(|X|-1)^2C\binom{n}{h}^{-1}$
        \item $\|\mathbb{E}[\hat{f_e}^{-\remvpt}(x,t) - f_e^{-\remvpt}(x,t)]\|_{\infty} \leq ln(16)|X|^2C\binom{n}{h}^{-1}$
    \end{enumerate}
\end{theorem}

Where $\hat{f_e}$ denotes the \codingensemble{} and $\hat{f_e}^{-\remvpt}$ denotes the \codingensemble{} with all models that have been trained on splits containing $\remvpt$ removed.

Therefore, using \codingensemble{}s we can assess the effects of removing one training point at a time. We note that instead of a single sample, we can group multiple samples together into a ``super sample'' if we suspect that these ``super samples'' will jointly play an important role before constructing the \codingensemble{}.   For example, we can group class specific training samples (e.g. airplane) to determine what role a given class plays in a given output.

We prespecify the hamming weights of the codes we assign to training samples to ensure the training samples are equally represented in the \codingensemble{}. It may also be desirable to ensure that after a given training sample is ablated from the \codingensemble{}, the remaining models still represent the rest of the training data equally. Unfortunately, to achieve this we would need to train $\mathcal{O}(|\trainset|)$ models as opposed to $\mathcal{O}(log(|\trainset|)^{1+\epsilon})$ models, which is intractable as $|\trainset|$ can be in the thousands or even billions \cite{schuhmann2022laion}. Formally:

\begin{theorem}
\label{thm-intersection}
    Let $E$ be some ground set, and let $M \subseteq 2^E\setminus \emptyset$ be some set system. Suppose there exist integers $z_1$ and $z_2$ such that following hold:
    \begin{enumerate}
        \item For all $e \in E$, $|\{m | m \in M, e \in m\}| = z_1$
        \item For all $e_1, e_2 \in E$, $|\{m | m \in M, e_1 \in m, e_2 \in m\}| = z_2$
    \end{enumerate}
    Then either $|M| \leq 1$, or $|M| \geq |E|$.
\end{theorem}

\subsection{We can approximate the counterfactual by computing a Jacobian}
\label{sec-approx-method}

Since unlearning can be performed instantaneously using \codingensemble{}s, the main bottleneck now becomes the generation of the counterfactual. This generation can be expensive, especially in generative models, and effectively intractable if we wish to evaluate these influences over an entire training dataset, for example to discover which part of the training data is most influential on a generated sample.

We address this by considering a linear approximation of the counterfactuals obtained via a Taylor expansion of the generative model. Let $\theta_1, \theta_2, ... \theta_n$ denote the parameters of the models of the \codingensemble{} $\hat{f_e}$, and let $v$ be a non-zero $n$-dimensional vector whose entries are all non-negative. We then define:

\begin{equation}
    \hat{f_e} \cdot v = \sum_{i=1}^n f(x, t, \theta_i) v_i
\end{equation}

If $u_0$ is the vector where all entries are $(1/n)$, then we have $\hat{f_e} \cdot u_0 = \hat{f_e}$. More generally, for any training point $\remvpt \in \trainset$, there exists a $u_{-\remvpt}$ such that $\hat{f_e} \cdot u_{-\remvpt} = \hat{f_e}^{-\remvpt}$. 

If we use $\hat{f_e} \cdot v$ as the denoiser instead of $\hat{f_e}$, then the generated sample is dependent on both the exogenous noise $\varepsilon$, and the value of $v$. To make this explicit, we define $y(v, \varepsilon)$ as the sample generated when $\varepsilon$ is provided as exogenous noise, and $\hat{f_e} \cdot v$ is used as the denoiser.
If we fix the exogenous noise, then we can perform the following first-order Taylor expansion around $u_0$:

\begin{equation}
    y(v, \varepsilon) = y(u_0, \varepsilon) + \frac{\partial y(x, \varepsilon)}{\partial x}\bigg|_{x = u_0} (v - u_0) + \mathcal{O}(\|v - u_0\|^2)
\end{equation}

$\frac{\partial y(x, \varepsilon)}{\partial x}$ is the Jacobian, which is an $m$-by-$n$ matrix, where $m$ is the dimension of the generated sample. This can be computed fairly efficiently by performing $n$ rounds of forward mode automatic differentiation. See Appendix \ref{appdx-training-experiments} for additional details.

The counterfactual can then be approximated by plugging in $u_{-\remvpt}$ into this approximation. Once the Jacobian is computed, evaluation is a straightforward matrix vector product, which is highly efficient and allows database scale evaluation. Since the most taxing part of evaluating the counterfactual is now computing the Jacobian, we will refer to this as the \emph{Jacobian approximation}.

\section{Results}

We first describe how we generally implement the concepts described in the previous section.
To train \acodingensemble{}, we first obtain the splits as described in Section \ref{sec-method-main}. Each model of the ensemble is then trained independently on a given split as described in Algorithm 1 in the paper by \citet{ho2020denoising}. Additional details regarding training can be found in Appendix \ref{appdx-training}.

Given \acodingensemble{} $\hat{f}_e$, we generate samples by first sampling isotropic Gaussian noise, then gradually denoising. We denoise by first predicting the noise using each model in the ensemble. We then average the predicted noise, remove the average predicted noise from the image, scale the image, and then add a small amount of isotropic Gaussian noise before repeating the denoising procedure. The full procedure is identical to the one described in Algorithm 2 in the paper by \citet{ho2020denoising}.

We define the exogenous noise involved in the process as both the isotropic Gaussian noise that we start with and the isotropic Gaussian noise that we add at each step. Therefore, fixing the exogenous noise fixes the output of the model, so a sample is completely determined by the exogenous noise used to generate it. This allows us to consider counterfactual denoising trajectories as long as we save the exogenous noise used to generate it.

To assess the effect of deleting a data point $\tilde{x}$ from the training set on a generated sample $y$, we remove every model that has been trained on a dataset that contains $\tilde{x}$, giving us a new \emph{ablated ensemble} which we denote by $\hat{f}_e^{-\tilde{x}}$. By Theorem \ref{thm-main}, there exists no other data point whose influence has been completely removed from the ensemble.

We then take the noise that was used to generate the sample $y$, and generate a new \emph{counterfactual sample} using that noise and the ablated model. We can then assess the difference between the counterfactual and original sample, either qualitatively or via some metric.

\subsection{Ablating classes of training data removes the \codingensemble{}'s ability to generate samples of those classes}
\label{sec-sanityCheck}

In this section we seek to establish that our approach of generating counterfactuals through temporary unlearning is viable. This is tricky, since it is difficult to establish ground truth for the influence of any given training sample, especially in the generative setting.

Therefore, we opt to first train \acodingensemble{} that encodes entire classes as opposed to individual training samples. We train \acodingensemble{} on the MNIST \cite{lecun2010mnist} dataset, which has clear and easily classifiable classes.
We achieve this by assigning the same code to all images belonging to the same class. Since the number of classes is small, we derive bit vectors from the Walsh matrix of size 8-by-8. We remove the row and column containing all 1s, and set the -1 entries to 0. This yields 7 unique bit vectors of length 7 where each vector has a hamming weight of 3, which we use as codes. Additionally, we also have the property that any pair of bit vectors have exactly one index where both vectors are 1.

We then assign the bit vectors to the classes 0, 1, 2, 3, 4, 6, and 8. The training images are then assigned to splits according to the bit vectors as described in Section \ref{sec-method-main}. Images are taken from the test split, and images belonging to classes 5, 7, and 9 are discarded. We then train 7 different diffusion models on the 7 splits. This gives us a controlled setting where we have ground truth.  If an entire class is ablated, then the ensemble's ability to generate samples from that class should be greatly diminished.

\begin{figure}

\hbox{
a) \includegraphics[width=0.6\textwidth]{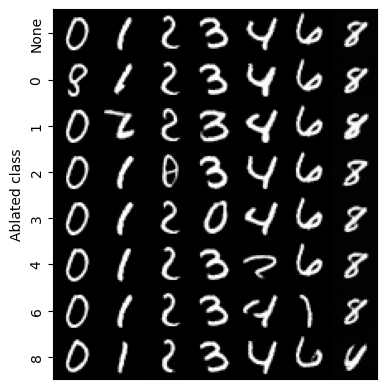}
\phantom{---}
\valign{#\cr
    b) \includegraphics[width=0.3\textwidth]{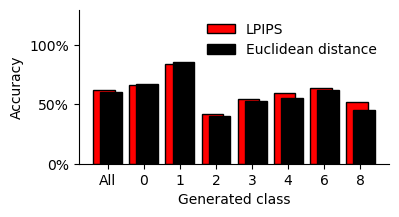}
    \vfill
    c) \includegraphics[width=0.3\textwidth]{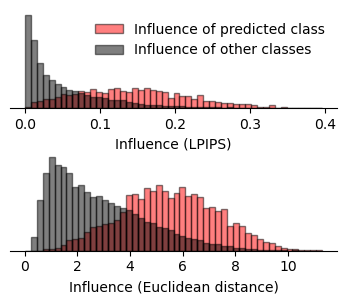}
    \vfill
    d) \includegraphics[width=0.3\textwidth]{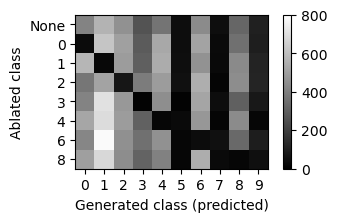}
    \cr}
}
 
  \caption{Ablating a class diminishes an \codingensemble{}'s ability to generate the ablated class}
  \label{fig-classShard}
  We demonstrate the validity of the ensemble ablation approach to assessing training data influence.
  
  a) Each column corresponds to a generated sample. For each generated sample, we generate 7 counterfactual samples, corresponding to removing classes. The row represents the counterfactual samples arising from a given model ablation. It is visually apparent which model has been ablated by looking at the rows. Examples here are cherry picked.

  b) We run the generated image through a CNN predictor, and we compute the Euclidean and Learned Perceptual Image Patch Similarity (LPIPS) distance between counterfactuals and their original image. We find that in general the counterfactual that differs most from the original image is the one in which the predicted class is ablated. The frequency of this is reported on the y-axis, and the x-axis is used to organize the images by class.

  c) We plot the distribution of LPIPS and Euclidean distances between counterfactual and original images. The red distribution contains the counterfactual where the predicted class of the original image is ablated, and the black distribution contains the rest of the counterfactuals. In both cases the red distribution is significantly greater than the black ($p \leq 10^{-300}$ via the Mann–Whitney U test).

  d) We plot the number of times an ablated ensemble generates a digit of a given class. Generated samples are organized in cells, where the row of the cell indexes the ablated ensemble that generated the sample, while the column indexes the class of the sample, as classified by a CNN predictor. As expected, if a class is ablated from an ensemble its ability to generate samples of that class is greatly diminished.
  Note that the model is not trained on 5, 7, and 9, so it produces very few of those digits.
\end{figure}

We then generate 3000 samples from the ensemble, along with 7 counterfactual samples for each sample, 1 for each class that gets ablated. A cherry-picked set of samples is shown in Figure \ref{fig-classShard}a, where it can be seen that ablating the class corresponding to the class of the original sample changes the sample much more significantly than ablating the other classes. We quantify this effect more rigorously by running the generated images through a CNN predictor to obtain their predicted classes, and observing how much effect ablating the predicted class of a generated sample has in contrast to ablating a different class. We quantify the effect using both Euclidean distance and Learned Perceptual Image Patch Similarity (LPIPS) \cite{zhang2018unreasonable} between the original sample and the counterfactual sample (Figure \ref{fig-classShard}b, c), and find that indeed ablating the predicted class tends to have a stronger effect on the counterfactual sample. The effect is sufficiently strong that we can predict the class of the image by looking at which class in the training set was the most influential with an accuracy of 62\% (LPIPS) and 61\% (Euclidean).
Finally, we looked at which digits are generated by the ensemble and its ablated counterparts, and find that when a class is ablated, the amount of samples of that class that are generated is greatly diminished (Figure \ref{fig-classShard}d).

These results show that ablating a training class strongly limits outputs of the ablated class, and establishes that our temporary unlearning technique erases knowledge of relevant parts the training set. We take care to note that we have not established that we do not overestimate influence. In fact, ablating a class different from the predicted class changes the class of the counterfactual sample 43\% of the time. This is less frequent than when the predicted class is ablated, which changes the class of the counterfactuals sample 93\% of the time. However, it is difficult to quantify how much of the 43\% is attributable to training influence and how much is attributable to other effects, such as reducing ensemble size (this effect is especially pronounced in this setting, since an ablated ensemble only contains 4 models)\footnote{We note that ensembles do appear to converge to a ``true'' ensemble average as ensemble size increases (see Figure \ref{fig-convergence}).}.

\subsection{Ensembles of diffusion models generate coherent output}

It is not immediately clear that ensembles of diffusion models should produce an outputs that are coherent, especially when the models were independently trained. While ensembling predictors can be straightforward via methods like averaging or majority voting \cite{sagi2018ensemble}, it is less clear that such approaches will work for generative models. Suppose we have a generative model $M: \mathbb{R}^n \rightarrow \mathbb{R}^m$ that maps $n$ dimensional isotropic Gaussian noise to $m$ dimensional outputs. Then if $A: \mathbb{R}^n \rightarrow \mathbb{R}^n$ is some unitary transformation, the composition $M \circ A$ defines the exact same distribution, yet may generate very different samples for identical input. A convex combination of such samples may not resemble samples from our desired distribution.

\begin{figure}

\begin{center}
    a) \includegraphics[width=0.9\textwidth]{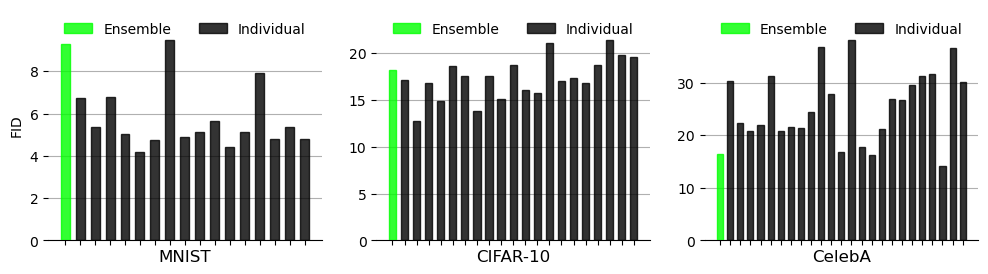}
    
    b) \includegraphics[width=0.9\textwidth]{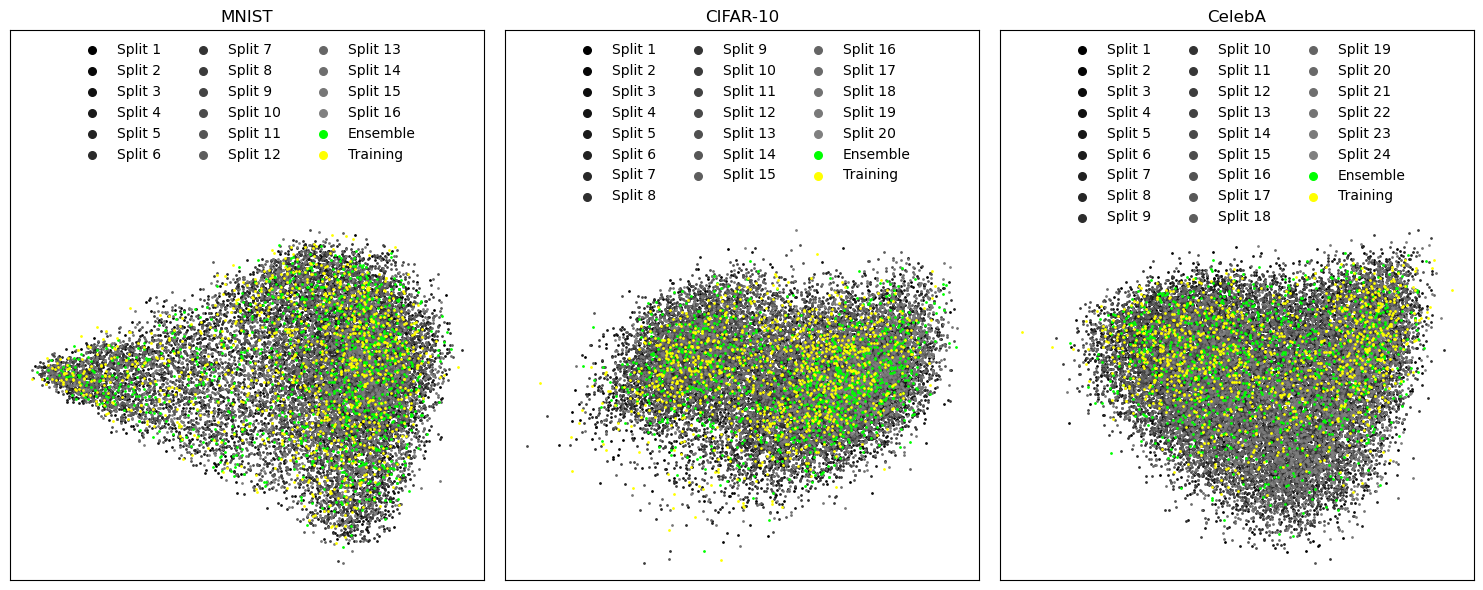}

    c) \includegraphics[width=0.3\textwidth]{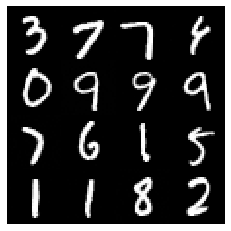}
    d) \includegraphics[width=0.3\textwidth]{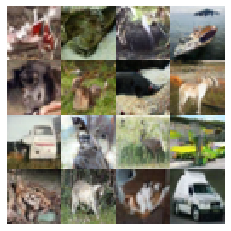}
    e) \includegraphics[width=0.3\textwidth]{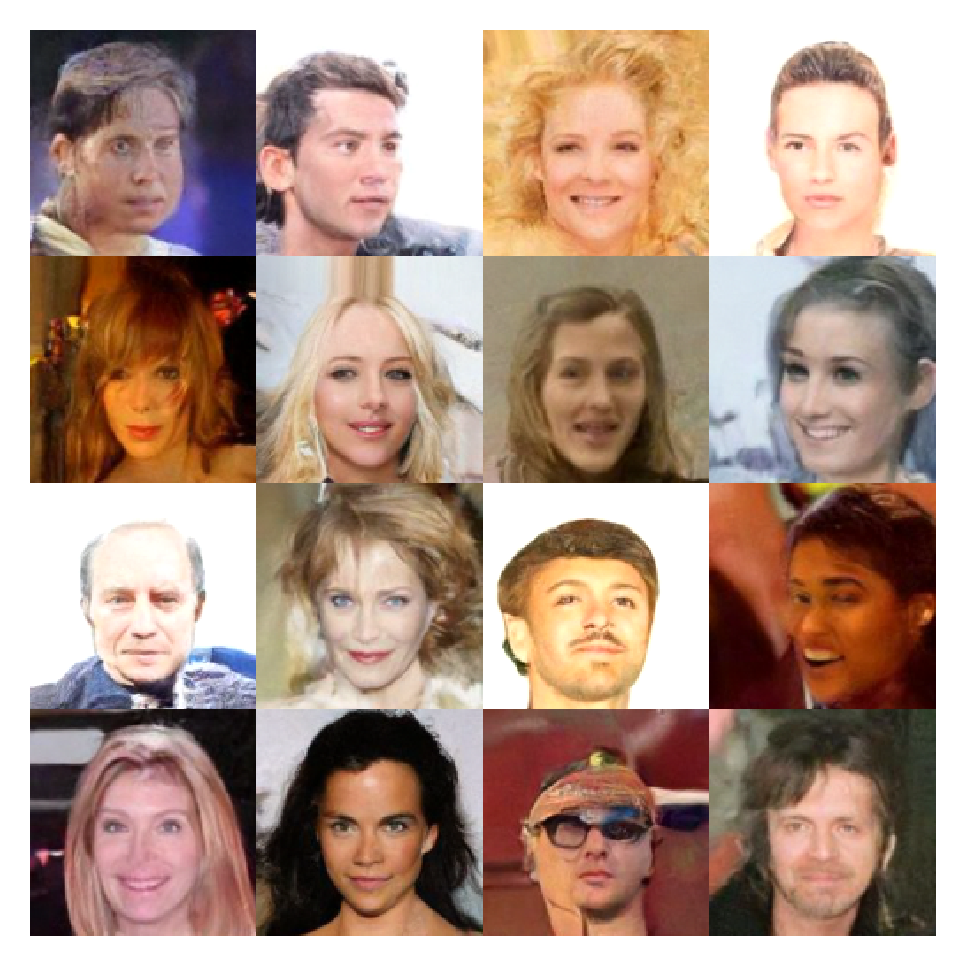}
\end{center}

\caption{Ensembles of diffusion models output samples that are comparable to individual models}
\label{fig-coherentOutput}

a) We compute the Frechet Inception Distance (FID) for the ensemble and each of its component models. We find that the FID of the ensemble is smaller than the FID of the component model with the largest FID. We report an FID of 9.29 for the MNIST \codingensemble{}, 18.20 for the CIFAR-10 \codingensemble{}, and 16.43 for the CelebA \codingensemble{}. We note that the FID reported for the CIFAR-10 ensemble is competitive with some of FID benchmarks reported by \cite{ho2020denoising} in their Table 1, although it is not state-of-the-art.

b) We plot the distribution of 1000 randomly selected samples from the samples generated by the ensemble, the samples generated by its component models, and the samples from the training set. Samples are given as feature vectors from the last layer of the Inception v3 \cite{DBLP:journals/corr/SzegedyVISW15} model and plotted in PCA space. Splits refer to individual models, of which there are 16 for MNIST, 20 for CIFAR-10, and 24 for CelebA.

c, d, e) Samples drawn from ensembles trained on MNIST (c), CIFAR-10 (d), and CelebA (e).
\end{figure}

Nonetheless, we find that ensembles of diffusion models do produce coherent output. First, we train three \codingensemble{}s: we train \acodingensemble{} of 16 models over 10000 28-by-28 MNIST images from the validation split \cite{lecun2010mnist}, \acodingensemble{} of 20 models over 60000 32-by-32 CIFAR-10 images \cite{krizhevsky2009learning}, and \acodingensemble{} of 24 models over 202599 128-by-128 CelebA images \cite{liu2015faceattributes}. We then generate 10000 samples from each distribution and evaluate their Frechet Inception Distance (FID) \cite{heusel2017gans} with respect to the training set. We do the same for each of the component models, which provides a baseline for the performance of unensembled models. The results are presented in Figure \ref{fig-coherentOutput}, and demonstrate that the output of the ensembled models are comparable to the output of the individual models.

\begin{figure}

\begin{center}
    a) \includegraphics[width=0.9\textwidth]{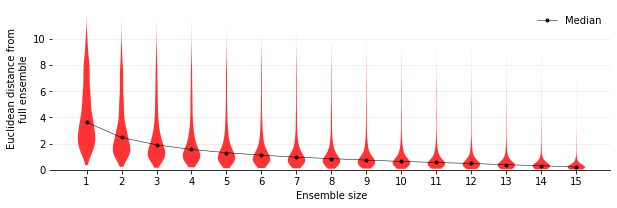}

    b) \includegraphics[width=0.9\textwidth]{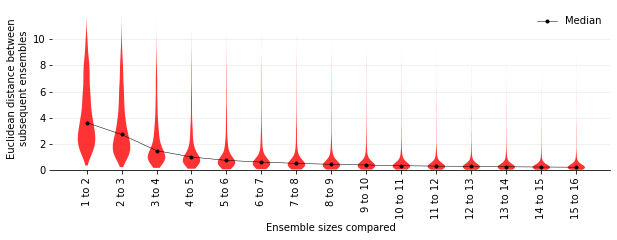}
    
    c) \includegraphics[width=0.9\textwidth]{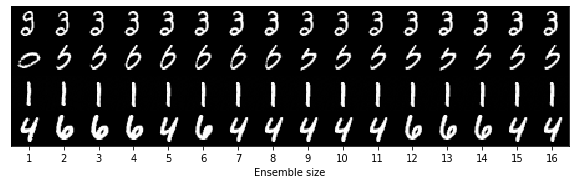}
\end{center}
 
  \caption{Ensembles of diffusion models converge as ensemble size increases}
  \label{fig-convergence}

  a) Given fixed exogenous noise, as we add more models to the ensemble, the final generated image gets increasingly closer to the image generated by the full ensemble. Distributions of Euclidean distances from the final image are given as violin plots, each violin representing 2752 samples generated from the \codingensemble{} trained on MNIST.

  b) The effect of adding a single model decreases with ensemble size. For example, the violin labelled ``5 to 6'' plots the distribution of Euclidean distances between images generated on an ensemble of 5 models and images generated on an ensemble of 6 models given the same exogenous noise.

  c) As more models are added to an ensemble we in general observe convergence.  We also occasionally observe alternating sequences such as in the last row.
\end{figure}

Finally, we find that ensemble outputs appear to converge to some ground truth output as ensemble sizes increase. We generate 2752 samples, where for each sample we also generate samples using the same exogenous noise using ensembles derived from a random sequence of nested subsets. The subsets are selected by selecting a permutation uniformly at random and creating a nested sequence by adding models to the set according to the order given by the permutation. The results are presented in
Figure \ref{fig-convergence}.

\subsection{Jacobian approximation of influence correlates with true influence}

\begin{figure}

\begin{center}
    a) \includegraphics[width=0.3\textwidth]{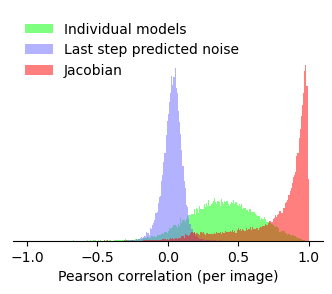}
    b) \includegraphics[width=0.3\textwidth]{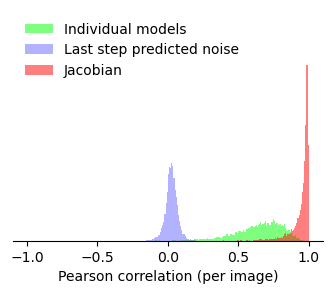}
    c) \includegraphics[width=0.3\textwidth]{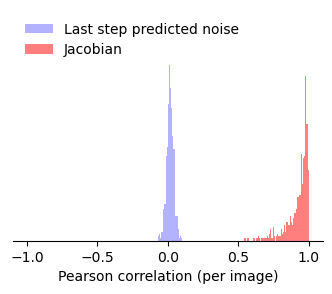}

    d) \includegraphics[width=0.3\textwidth]{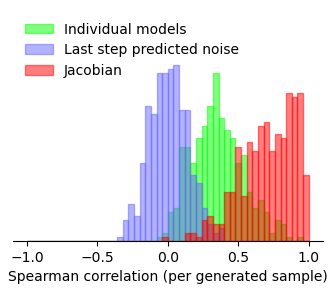}
    e) \includegraphics[width=0.3\textwidth]{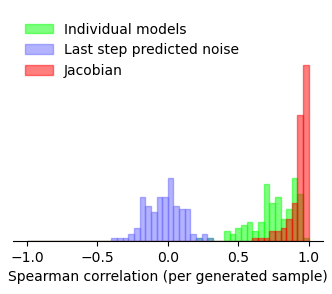}
    f) \includegraphics[width=0.3\textwidth]{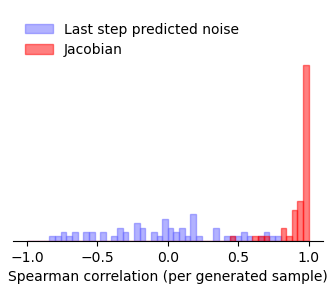}
\end{center}

  \caption{The Jacobian approximates counterfactual samples}
  \label{fig-jacobian}
  
  The Jacobian approximation is superior to baseline controls for MNIST (a), CIFAR-10 (b), and CelebA (c).  The per image Pearson correlations of the differences between the true counterfactual and Jacobian approximation counterfactual are significantly higher than those of the differences between the true counterfactuals and the baseline appoximation counterfactuals ($p \leq 10^{-168}$ via a Mann–Whitney U test in all cases).  The Jacobian approximation is also superior to baseline controls at identifying counterfactuals with the largest difference from the original image for MNIST (d), CFAR-10 (e), and CelebA (f).   Actual counterfactuals associated with each sample (382 samples, 64 counterfacturals each for MNIST; 128 samples, 64 counterfactuals each for CFAR-10; 64 samples, 8 counterfactuals each for CelebA) are compared to their Jacobian approximations and baseline controls via Spearman rank correlation.  Jacobian approximation rank correlations are significantly higher than rank correlations for baseline approximations by the Mann–Whitney U test ($p\leq10^{-80}$ for any pair in (d), $p\leq10^{-26}$ for any pair in (e), and $p\leq 10^{-21}$ for the distributions in (f)).

\end{figure}

Next, we take our MNIST, CIFAR-10, and CelebA \codingensemble{}s and generate samples along with their counterfactuals and Jacobians. For the MNIST \codingensemble{} we generate 382 samples with 64 counterfactuals for each sample, for the CIFAR-10 \codingensemble{} we generate 128 samples with 64 counterfactuals for each sample, and for the CelebA \codingensemble{} we generate 64 samples with 8 counterfactuals for each sample. Then using Jacobians of each generated sample we compute the approximate counterfactuals as described in Section \ref{sec-approx-method}.
We find that the approximate counterfactuals are able to approximate the changes in pixel intensities on a per image basis (Figure \ref{fig-jacobian}a, \ref{fig-jacobian}b, \ref{fig-jacobian}c), and are able to capture the magnitudes of the changes as well (Figure \ref{fig-jacobian}d, \ref{fig-jacobian}e, \ref{fig-jacobian}f).

We compare the Jacobian approximation against two baselines.  In the ``Last step predicted noise'' baseline, we generate the image normally until the final denoising step, where we drop the outputs of all models that we wish to ablate.  For the ``Individual models'' baseline, we generate independently from individual models using the same exogenous noise. We then subtract the original image from each of the individually generated images, resulting in a set of residuals. We negate the residuals corresponding to models that were ablated and average the residuals, which we then add to the original image.  We find that the Jacobian approximation significantly outperforms both these baselines in the MNIST and CIFAR-10 \codingensemble{}s. For CelebA we opted to only compare against the ``Last step predicted noise'' baseline, which the Jacobian approximation significantly outperforms. These comparisons are presented in Figure \ref{fig-jacobian}.

\begin{figure}

\begin{center}
    a) \includegraphics[width=0.45\textwidth]{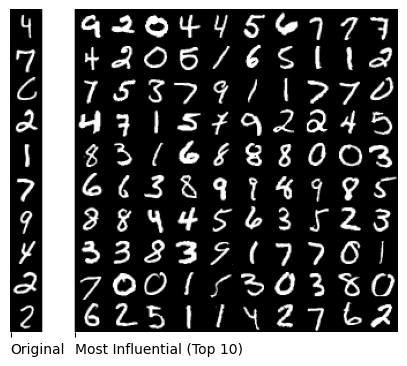}
    b) \includegraphics[width=0.45\textwidth]{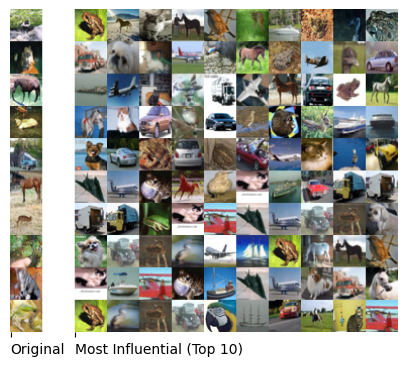}
    
    c) \includegraphics[width=0.45\textwidth]{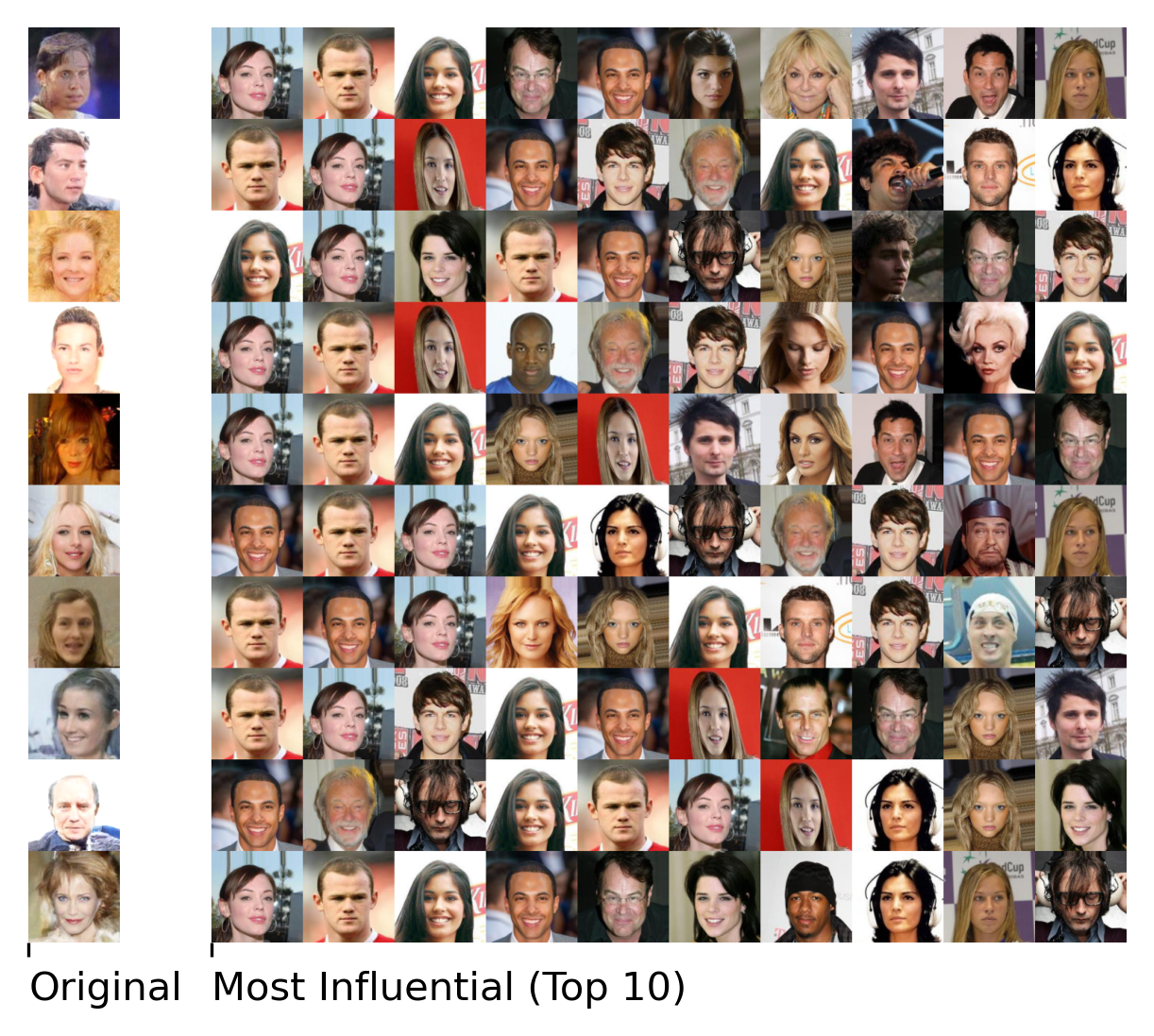}
    d) \includegraphics[width=0.45\textwidth]{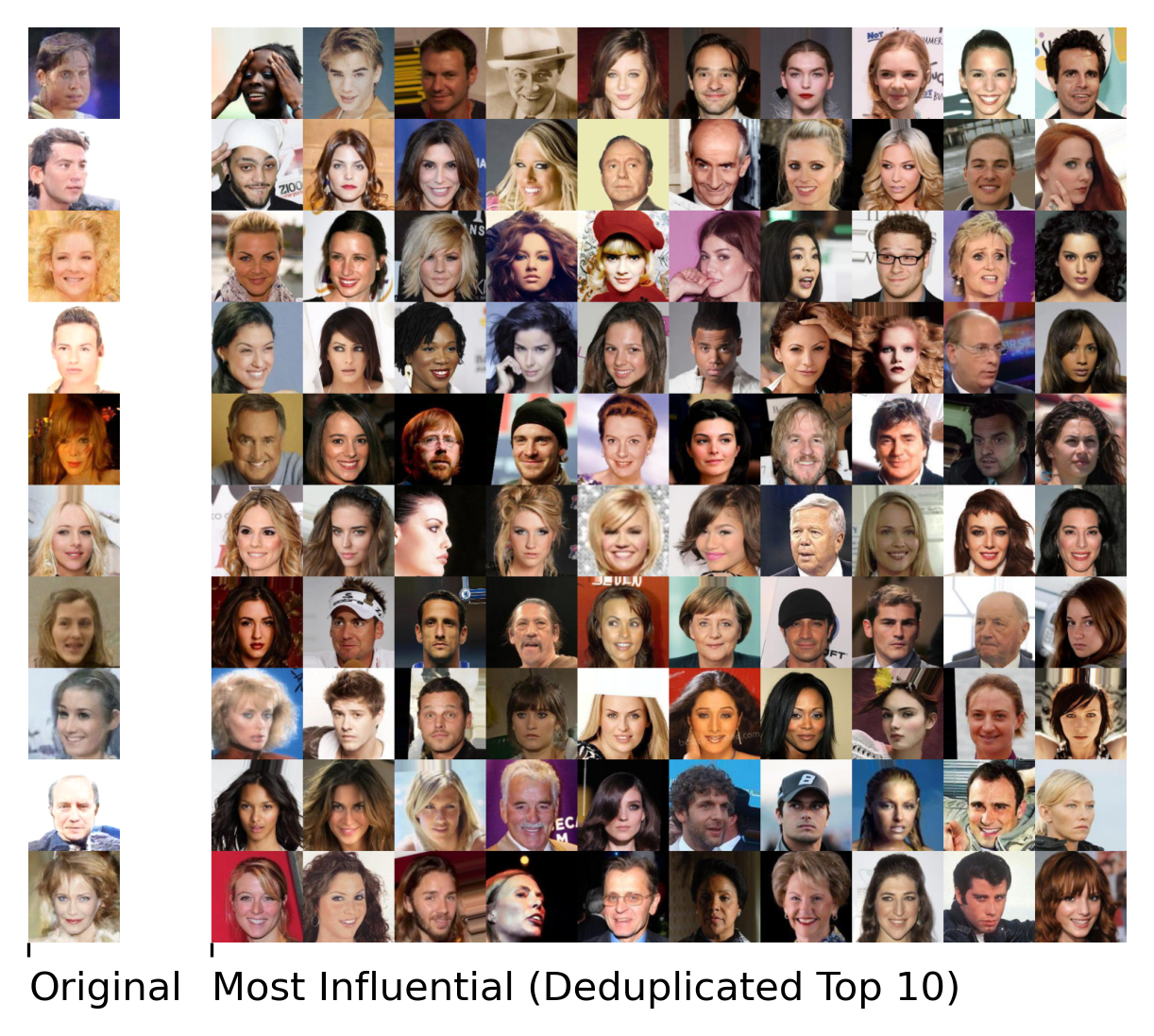}
\end{center}
 
  \caption{The Jacobian approximation enables database scale search for influential samples}
  \label{fig-databaseSearch}

  We find the 10 most influential training images as approximated by the Jacobian approximation across entire training sets for samples generated by \codingensemble{}s trained on a) MNIST (10000 images), b) CIFAR-10 (60000 images), and c) CelebA (202599 images). CelebA in particular appears to contain many ``generally influential'' images, so we also give the 10 most influential training images while attempting to discount these ``generally influential'' images in d). We do this by removing any image that appears in more than one of the 10 top 10 lists and iterating until no image appears twice.  Each row corresponds to a generated sample, with the generated sample on the left and the most influential training images given on the right.
\end{figure}

As noted in Section \ref{sec-approx-method}, while the evaluation of the Jacobian can be computationally intensive, once computed we will be able to evaluate the approximate counterfactual with a matrix vector product. We therefore compute for each image in the training set their approximate influence on each generated sample (measured as the Euclidean distance from the approximate counterfactual to the original sample). We present the top 10 most influential training images on a random selection of generated samples in Figure \ref{fig-databaseSearch}. We note that it is not immediately clear why the presented images would be the most influential, and it seems that some images are simply ``generally influential'' since they appear as highly influential in multiple samples. This is in line with previous observations that unconditioned diffusion models trained on sufficiently large corpuses of data, such as those we train here, do not appear to exhibit significant copying from their datasets \cite{somepalli2023diffusion}.

\section{Discussion}

We presented a viable method of assessing the influence of training data on samples generated by diffusion models by tracing influence through \codingensemble{}s. We showed that ensembles of diffusion models output intelligible samples, and that ablating specific \codingensemble{} members allows us to identify the influence of specific training data. Finally, we presented a method for approximating this influence by computing the Jacobian, enabling the search of influential training data on database scales.

\subsection{Limitations and Future work}

One of the key limitations of our method is the need to train and sample an ensemble of models. While the required size of the ensemble is approximately logarithmic in the size of the training data, this still represents a possible 10-50 fold increase in the resources required for training and sampling. It also requires designing a diffusion model from scratch with training data attributability in mind.  

As mentioned before, our ground truth for training data influence involves model ablation that removes training data. While we have shown in Section \ref{sec-sanityCheck} that ablation is a viable measure of influence, we can only guarantee that the resulting model has not seen the removed training sample. We cannot guarantee that some other effect does not come into play.  %
Designing improved schemes for assigning codes that minimize unintended ablation effects also remains the subject of future work.

We have applied our approach to unconditioned pixel space models, and observed that the influence of training data can appear unintuitive. However, it has been previously shown that in the case of text conditioned latent space diffusion models, we can find clear instances of copying \cite{somepalli2023diffusion}. We intend to extend our work to such models.

Since our approach evaluates training data attribution for a single generated sample at a time it does not assess the global influence of training data over the the distribution of generative model outputs. It is even possible that the unlearning defines a non-identity yet measure preserving map. It could completely change a given generated sample, yet the probability of obtaining the original sample remains unchanged\footnote{For example, a rotation on a Euclidean space with isotropic Gaussian measure.}.

\subsection{Statement on Broader Impact}

As stated in our introduction, the increasingly wide adoption of diffusion models in various domains means that our work will have wide-ranging implications across those domains. This work can be applied to improve the interpretability of diffusion models, enabling practitioners to identify and mitigate biases or unintended consequences introduced by the training data. Understanding how training data influences model outputs is crucial in domains such as therapeutics and law, where generative models may take on important roles. 

The widespread adoption of diffusion models for creative uses has also raised serious ethical and legal concerns surrounding the use of training data. While yet unresolved, we expect the ability to trace model generations to training data will be an important consideration in these discussions and play a key role in its resolution.

\section*{Data availability}
Code and data for this paper can be found at \url{https://github.com/zheng-dai/GenEns}.

\newpage
\appendix
\section*{Appendix}
\section{Proofs}
\label{appdx-proofs}
\subsection{Theorem \ref{thm-main}}

Recall that given a dataset $X$, we assign each $x \in X$ a unique bit vector of length $n$ with Hamming weight $h$. We then create $n$ splits of the original dataset such that $x$ belongs to split $i$ if and only if the $i$th entry in its bit vector is 1.

First, we show that if we remove all models that have been trained on $x$, there exists no other $x' \in X$ such that all models trained on $x'$ is also removed. This is true because each bit vector has the same Hamming weight, so it is impossible for the models trained on $x'$ to be a subset of the models trained on $x$.

Next, we show that if the bit vectors are assigned uniformly at random, and additionally $\binom{n}{h} \geq 2|X|$, $n = 2h$ (note that implicitly $n \geq 2$), and $\linf{f(x',t',\theta')} \leq C$ for all inputs $x'$, $t'$, and $\theta'$ we have for arbitrary $\tilde{x} \in X$::

\begin{enumerate}
        \item $\|\mathbb{E}[\hat{f_e}(x,t) - f_e(x,t)]\|_{\infty} \leq ln(16)(|X|-1)^2C\binom{n}{h}^{-1}$
        \item $\|\mathbb{E}[\hat{f_e}^{-\tilde{x}}(x,t) - f_e^{-\tilde{x}}(x,t)]\|_{\infty} \leq ln(16)|X|^2C\binom{n}{h}^{-1}$
\end{enumerate}

\subsubsection{Statement 1}

To make the following derivation more explicit, let $\mathcal{R}$ denote the distribution of random input that is used for training. Let $X$ be denote the training set, and let $U$ denote the uniform distribution over $2^X$. Let $BV(n,h)$ denote set of bit vectors of length $n$ with Hamming weight $h$, and let $W$ denote the distribution obtained with $|X|$ samples are sampled sequentially and uniformly from $BV(n,h)$ without replacement. Let $W'$ denote the distribution obtained with $|X|$ samples are sampled sequentially and uniformly from $BV(n,h)$ with replacement. Let $Z$ denote the set of events that are in the support of $W'$ but not in the support of $W$. Given some sequence $S$ drawn from $W$, let $D_i(S) \in 2^X$ denote a subset of $X$ where, given some total order over $X$, includes the $j$th element of $X$ if and only if the $j$th bit vector of $S$ has a 1 at position $i$.

Then we have:

\begin{align}
    \mathbb{E}[\hat{f_e}(x,t) - f_e(x,t)] &=
    \mathbb{E}_{V \sim W, r \sim \mathcal{R}}[\frac{1}{n}\sum_{i=1}^n f(x,t,\mathcal{A}(D_i(V),r))]
    -
    \mathbb{E}_{S \sim U, r \sim \mathcal{R}}[f(x,t,\mathcal{A}(S,r))]
    \\
    &= \frac{1}{n}\sum_{i=1}^n
    \mathbb{E}_{V \sim W, r \sim \mathcal{R}}[f(x,t,\mathcal{A}(D_i(V),r))]
    -
    \mathbb{E}_{S \sim U, r \sim \mathcal{R}}[f(x,t,\mathcal{A}(S,r))]
    \\
    &= 
    \mathbb{E}_{V \sim W, r \sim \mathcal{R}}[f(x,t,\mathcal{A}(D_1(V),r))]
    -
    \mathbb{E}_{S \sim U, r \sim \mathcal{R}}[f(x,t,\mathcal{A}(S,r))]
\end{align}

Which holds because $\mathbb{E}_{V \sim W, r \sim \mathcal{R}}[f(x,t,\mathcal{A}(D_1(V),r)] = \mathbb{E}_{V \sim W, r \sim \mathcal{R}}[f(x,t,\mathcal{A}(D_2(V),r)]$ for any $i$ and $j$ due to symmetry. We then have:

\begin{align}
    \mathbb{E}_{V \sim W, r \sim \mathcal{R}}[f(x,t,\mathcal{A}(D_1(V),r))]
    &=
    \mathbb{E}_{V \sim W', r \sim \mathcal{R}}[f(x,t,\mathcal{A}(D_1(V),r)) | V \notin Z]
    \label{eqn-8}
\end{align}

The is because one procedure for uniquely assigning elements of $BV(n,h)$ uniformly at random is via rejection sampling: first sample an assignment of bit vectors uniformly at random without consideration for whether the each data point is assigned a unique bit vector, then reject and resample if there is a collision. We also have:

\begin{align}
    \mathbb{E}_{S \sim U, r \sim \mathcal{R}}[f(x,t,\mathcal{A}(S,r))]
    =&
    \mathbb{E}_{V \sim W', r \sim \mathcal{R}}[f(x,t,\mathcal{A}(D_1(V),r))]
    \\=&
    \mathbb{E}_{V \sim W', r \sim \mathcal{R}}[f(x,t,\mathcal{A}(D_1(V),r)) | V \in Z]\prob{Z \ni V \sim W'}
    +\notag\\
    &\mathbb{E}_{V \sim W', r \sim \mathcal{R}}[f(x,t,\mathcal{A}(D_1(V),r)) | V \notin Z]
    \prob{Z \not\ni V \sim W'}
    \label{eqn-11}
\end{align}

Because half the elements of $BV(n,h)$ have 0 at coordinate 1 and half have 1 at coordinate 1 (since $n = 2h$), so sampling uniformly from $BV(n,h)$ and only considering the first coordinate is equivalent to flipping a fair coin.

Combining Equation \ref{eqn-8} with Equation \ref{eqn-11} gives:

\begin{align}
    \|\mathbb{E}[\hat{f_e}(x,t) - f_e(x,t)]\|_{\infty} = 
    \|
    \bigg(&\mathbb{E}_{V \sim W', r \sim \mathcal{R}}[f(x,t,\mathcal{A}(D_1(V),r) | V \notin Z]
    \notag\\&-
    \mathbb{E}_{V \sim W', r \sim \mathcal{R}}[f(x,t,\mathcal{A}(D_1(V),r)) | V \in Z]
    \bigg)
    \prob{Z \ni V \sim W'}
    \|_{\infty}\\
    \leq 2&C\prob{Z \ni V \sim W'}
    \label{eqn-14}
\end{align}

Since each coordinate in the output of $f$ is bounded between $-C$ to $C$. $\prob{Z \ni V \sim W'}$ is the problem of obtaining a collision when drawing with replacement $|X|$ times out of $\binom{n}{h}$ elements, which we can estimate loosely as the following:

\begin{align}
    \prob{Z \ni V \sim W'} &= 1 - \frac{ \binom{n}{h}! }{ (\binom{n}{h} - |X|)! \binom{n}{h}^{|X|} }
    \label{eqn-13}
    \\
    &\leq 1 - (1-\frac{|X|-1}{\binom{n}{h}})^{|X|-1}
    \\
    &\leq 1 - \bigg((1-\frac{1}{(|X|-1)^{-1}\binom{n}{h}})^{(|X|-1)^{-1}\binom{n}{h}}\bigg)^{(|X|-1)^2\binom{n}{h}^{-1}}
    \\
    &\leq 1 - e^{-ln(4)(|X|-1)^2\binom{n}{h}^{-1}}
    \label{eqn-19}
    \\
    &\leq ln(4)(|X|-1)^2\binom{n}{h}^{-1}
    \label{eqn-20}
\end{align}

Equation \ref{eqn-19} holds since $\binom{n}{h} \geq 2|X|$. Plugging Equation \ref{eqn-20} into Equation \ref{eqn-14} yields the desired statement:

\begin{equation}
    \|\mathbb{E}[\hat{f_e}(x,t) - f_e(x,t)]\|_{\infty} \leq
    ln(16)(|X|-1)^2C\binom{n}{h}^{-1}
\end{equation}

\subsubsection{Statement 2}

To prove the second inequality, without loss of generality we let $\tilde{x}$ be the first element in the enumeration of the dataset $X$. Let $Q$ denote the event that $\tilde{x}$ is assigned the bit vector whose first half consists entirely of 1s and second half consists entirely of 0s. Then:

\begin{align}
    \mathbb{E}[\hat{f_e}^{-\tilde{x}}(x,t)] &=
    \mathbb{E}_{V \sim W, r \sim \mathcal{R}}[\frac{1}{n-h}\sum_{i=h+1}^n f(x,t,\mathcal{A}(D_i(V),r))
    |
    Q]\\
    &=
    \mathbb{E}_{V \sim W, r \sim \mathcal{R}}[f(x,t,\mathcal{A}(D_n(V),r))
    |
    Q]
    \label{eqn-20b}
\end{align}

This is due to symmetry: if the bit vector were permuted the conditioned expectation would still be equal.

Let $U'$ be the uniform distribution over $2^{X\setminus\tilde{x}}$. We also have:

\begin{align}
    f_e^{-\tilde{x}}(x,t)
    = \mathbb{E}_{S \sim U', r \sim \mathcal{R}}[f(x,t,\mathcal{A}(S,r))]
    \label{eqn-21}
\end{align}

Let $B$ be a distribution over vectors of length $|X|-1$ such that samples drawn from it have each entry drawn uniformly and independently from the uniform distribution between 0 and 1. Let $G: [0,1]^{|X|-1} \times [0,1] \rightarrow 2^{X\setminus\tilde{x}}$ be a function that takes in two values: the first is some sample $b \sim B$, and the second is a real value between 0 and 1. Then let $G(b,p)$ be the set that contains the $i$th element of $X\setminus\tilde{x}$ (given some ordering over that set) if and only if the $i$th coordinate of $b$ is at least $p$.

Then the distribution of $G(b, 0.5)$ where $b \sim B$ is identical to the distribution of $U'$. It is also the case that the distribution of $G(b, (\binom{n}{h}/2)(\binom{n}{h} - 1)^{-1})$ is equal to the distribution of $D_n(V)$, where $V \sim W'$ conditioned on $Q$ (denoted by $W'|Q$). For simplicity we will denote $(\binom{n}{h}/2)(\binom{n}{h} - 1)^{-1})$ as $b'$.

We then have:

\begin{align}
    \mathbb{E}[f_e^{-\tilde{x}}(x,t)]
    =& \mathbb{E}_{b \sim B, r \sim \mathcal{R}}[f(x,t,\mathcal{A}(G(b,0.5),r))]
    \\
    =& \mathbb{E}_{b \sim B, r \sim \mathcal{R}}[f(x,t,\mathcal{A}(G(b,b'),r))] +
    \bigg(
    \mathbb{E}_{b \sim B, r \sim \mathcal{R}}[f(x,t,\mathcal{A}(G(b,0.5),r))]
    \notag\\&-
    \mathbb{E}_{b \sim B, r \sim \mathcal{R}}[f(x,t,\mathcal{A}(G(b,b'),r))]
    \bigg)
    \label{eqn-23}
\end{align}

Focusing on the first term, we have:

\begin{align}
    \mathbb{E}_{S \sim G', r \sim \mathcal{R}}[f(x,t,\mathcal{A}(S,r))]
    =&
    \mathbb{E}_{V \sim W', r \sim \mathcal{R}}[f(x,t,\mathcal{A}(D_n(V),r)) | Q]\\
    =&
    \mathbb{E}_{V \sim W', r \sim \mathcal{R}}[f(x,t,\mathcal{A}(D_n(V),r))
    | Q, V \in Z]\prob{Z \ni V \sim W'|Q}
    \notag\\&+
    \mathbb{E}_{V \sim W', r \sim \mathcal{R}}[f(x,t,\mathcal{A}(D_n(V),r))
    | Q, V \notin Z]\prob{Z \not\ni V \sim W'|Q}
    \\
    =&
    \mathbb{E}_{V \sim W', r \sim \mathcal{R}}[f(x,t,\mathcal{A}(D_n(V),r))
    | Q, V \in Z]\prob{Z \ni V \sim W'|Q}
    \notag\\&-
    \mathbb{E}_{V \sim W', r \sim \mathcal{R}}[f(x,t,\mathcal{A}(D_n(V),r))
    | Q, V \notin Z]\prob{Z \ni V \sim W'|Q}
    \notag\\&+
    \mathbb{E}[\hat{f_e}^{-\tilde{x}}(x,t)]
\end{align}

The last line follows from Equation \ref{eqn-20b}.

Plugging this into Equation \ref{eqn-23}, then plugging that into Equation \ref{eqn-21} and rearranging yields:

\begin{align}
    \mathbb{E}[\hat{f_e}^{-\tilde{x}}(x,t) - f_e^{-\tilde{x}}(x,t)]
    =&
    \prob{Z \ni V \sim W'|Q}
    \bigg(
    \notag\\&
    \mathbb{E}_{V \sim W', r \sim \mathcal{R}}[f(x,t,\mathcal{A}(D_n(V),r))
    | Q, V \notin Z]
    \notag\\&
    -
    \mathbb{E}_{V \sim W', r \sim \mathcal{R}}[f(x,t,\mathcal{A}(D_n(V),r))
    | Q, V \in Z]
    \bigg)
    \notag\\&
    +
    \bigg(
    \mathbb{E}_{b \sim B, r \sim \mathcal{R}}[f(x,t,\mathcal{A}(G(b,b'),r))]
    \notag\\&-
    \mathbb{E}_{b \sim B, r \sim \mathcal{R}}[f(x,t,\mathcal{A}(G(b,0.5),r))]
    \bigg)
\end{align}

Taking the norm gives us:

\begin{align}
    \|\mathbb{E}[\hat{f_e}^{-\tilde{x}}(x,t) - f_e^{-\tilde{x}}(x,t)]\|_{\infty}
    \leq&
    \|\prob{Z \ni V \sim W'|Q}
    \bigg(
    \notag\\&
    \mathbb{E}_{V \sim W', r \sim \mathcal{R}}[f(x,t,\mathcal{A}(D_n(V),r))
    | Q, V \notin Z]
    \notag\\&
    -
    \mathbb{E}_{V \sim W', r \sim \mathcal{R}}[f(x,t,\mathcal{A}(D_n(V),r))
    | Q, V \in Z]
    \bigg)\|_{\infty}
    \notag\\&
    +\|
    \bigg(
    \mathbb{E}_{b \sim B, r \sim \mathcal{R}}[f(x,t,\mathcal{A}(G(b,b'),r))]
    \notag\\&-
    \mathbb{E}_{b \sim B, r \sim \mathcal{R}}[f(x,t,\mathcal{A}(G(b,0.5),r))]
    \bigg)\|_{\infty}
    \\
    \leq&
    \prob{Z \ni V \sim W'|Q}2C
    \notag\\&
    +
    \| \mathbb{E}_{b \sim B, r \sim \mathcal{R}}[f(x,t,\mathcal{A}(G(b,0.5,r)) - 
    f(x,t,\mathcal{A}(G(b,b'),r))] \|_{\infty}
    \\
    \leq&
    \prob{Z \ni V \sim W'|Q}2C
    +
    \prob{G(b,0.5) \neq G(b,b'), b \sim B}2C
    \\
    =&
    2C\bigg(
    \prob{Z \ni V \sim W'|Q}
    +
    \prob{G(b,0.5) \neq G(b,b'), b \sim B}
    \bigg)
\end{align}

Where once again we rely on the fact that each entry of the output of $f$ is bound between $-C$ and $C$.

For the first term, we note that conditioning on $Q$ (i.e. what the first draw is) does not change the probability of a collision, so it is simply $ln(4)(|X|-1)^2\binom{n}{h}^{-1}$ as shown in the derivation starting from Equation \ref{eqn-13} to Equation \ref{eqn-20}. For the second term, it is simply the probability that after $|X|-1$ draws from the uniform distribution, none of them falls into $(0.5, b')$. The probability that a single draw falls in is:

\begin{align}
    (\binom{n}{h}/2)(\binom{n}{h}-1)^{-1} - 0.5 &=
(\binom{n}{h}/2)(\binom{n}{h}-1)^{-1} - (\binom{n}{h}/2)\binom{n}{h}^{-1}\\
&=(\binom{n}{h}/2)\frac{\binom{n}{h} - (\binom{n}{h}-1)}{\binom{n}{h}(\binom{n}{h}-1)}\\
&= \frac{1}{2(\binom{n}{h} - 1)}\\
&\leq \frac{1}{\binom{n}{h}}
\end{align}

The last relation holds as long as $\binom{n}{h} \geq 2$. Then the probability that at least one of the $|X|-1$ draws falls in is:

\begin{align}
    1 - (1-\frac{1}{\binom{n}{h}})^{|X|-1} &=
    1 - \bigg(
    (1-\frac{1}{\binom{n}{h}})^{\binom{n}{h}}
    \bigg)
    ^{(|X|-1)\binom{n}{h}^{-1}}
    \\
    &\leq 1 - e^{-ln(4)(|X|-1)\binom{n}{h}^{-1}}
    \\
    &\leq ln(4)(|X|-1)\binom{n}{h}^{-1}
\end{align}

So altogether we have:

\begin{align}
    \prob{Z \ni V \sim W'|Q}
    +
    \prob{G(b,0.5) \neq G(b,b'), b \sim B}
    &\leq
    ln(4)(|X|-1)\binom{n}{h}^{-1}
    +
    ln(4)(|X|-1)^2\binom{n}{h}^{-1}\\
    &\leq
    ln(4)|X|^2\binom{n}{h}^{-1}
\end{align}

Which yields the desired statement:

\begin{equation}
    \|\mathbb{E}[\hat{f_e}^{-\tilde{x}}(x,t) - f_e^{-\tilde{x}}(x,t)]\|_{\infty} \leq ln(16)|X|^2C\binom{n}{h}^{-1}
\end{equation}

\subsection{Theorem \ref{thm-intersection}}

Let $E$ be some ground set, and let $M \subseteq 2^E\setminus \emptyset$. Suppose there exist integers $z_1$ and $z_2$ such that following hold:
    \begin{enumerate}
        \item For all $e \in E$, $|\{m | m \in M, e \in m\}| = z_1$
        \item For all $e_1, e_2 \in E$, $|\{m | m \in M, e_1 \in m, e_2 \in m\}| = z_2$
    \end{enumerate}

Our goal is to then show that either $|M| \leq 1$, or $|M| \geq |E|$. If $|M| \leq 1$, then both conditions are trivially satisfied, so we suppose that $|M| \geq 1$, and show that $|M|$ must be at least $|E|$ or greater.

Let us impose a total ordering over the elements of $E$. Then each element of $2^E$ can be expressed as a bit vector of length $|E|$, where the $i$th element is 1 if and only if the $i$th element of $E$ belongs to the set that the bit vector represents. Let $M'$ be a matrix whose columns are the vectors representing the elements of $M$. The two conditions can then be restated as the following equivalent conditions:

\begin{enumerate}
    \item The sum of each row of $M'$ is equal to $z_1$
    \item The dot product between any two distinct rows of $M'$ is $z_2$
\end{enumerate}

Since the vectors are bit vectors, the second condition implies that the Euclidean distance between any two distinct rows is $\sqrt{2z_1 - z_2}$, so the rows are all mutually equidistant in Euclidean space. Furthermore, since all the rows sum to $z_1$, they all reside on a hyperplane, where the normal is the vector where all entries are $\sqrt{1/|M|}$.

Suppose for the sake of contradiction that $|M| < |E|$. Then that means there are $|E|$ points that lie on an $|M|-1$ dimensional hyperplane that are all mutually equidistant. This implies that there is a way of arranging $|E|$ points in an Euclidean space of dimension $|M|-1$ such that they are mutually equidistant.

The $|E|$ rows must all be distinct, because if any pair are not, that implies that their dot product is $z_1$. Since all rows share the dot product, that implies that all rows are identical. But then that would imply that the elements of $M$ are either $E$ or the empty set. The empty set cannot belong to $M$, so $M = \{E\}$. But then $|M| = 1$, and we had supposed that $|M| > 1$.

Therefore, there should be a way of arranging $|E|$ distinct points in an Euclidean space of dimension $|M|-1$, where $|M| < |E|$. However, this is impossible, since given an $n$ dimensional Euclidean space the largest number of distinct mutually equidistant points we can have is $n+1$. Thus, by contradiction we have $|M| \geq |E|$, as desired.

\section{Training Details}
\label{appdx-training}

\subsection{Data processing}

In this work we make use of the MNIST, CIFAR-10, and CelebA datasets. The channel values of the images were scaled and shifted so that all values lie between -1 and 1. We used images from the test split of MNIST (10000 images total), and we combined both splits for CIFAR-10 (60000 images total) and CelebA (202599 images total). CelebA images were center cropped and resized to 128-by-128 (resizing was performed via torchvision.transforms.Resize on PIL images with default parameters). We split MNIST in two ways: one for the class specific splits and one for the main experiment.

For the class specific splits, we generated 7 splits based on the third Walsh matrix:

\begin{equation}
\begin{bmatrix}
1 & 1 & 1 & 1 & 1 & 1 & 1 & 1\\
1 &-1 & 1 &-1 & 1 &-1 & 1 &-1\\
1 & 1 &-1 &-1 & 1 & 1 &-1 &-1\\
1 &-1 &-1 & 1 & 1 &-1 &-1 & 1\\
1 & 1 & 1 & 1 &-1 &-1 &-1 &-1\\
1 &-1 & 1 &-1 &-1 & 1 &-1 & 1\\
1 & 1 &-1 &-1 &-1 &-1 & 1 & 1\\
1 &-1 &-1 & 1 &-1 & 1 & 1 &-1
\end{bmatrix}    
\end{equation}

We can convert this to a bit matrix by turning all the $-1$ values to $0$. If we remove the first row and column, the set of rows (or columns) are a set of 7 bit vectors all with Hamming weights of 3, and whose pairwise dot products are all equal.

We then assign the bit vectors to the classes 0, 1, 2, 3, 4, 6, and 8 in that order. All images belonging to a specific class are included in dataset $i$ if and only if the $i$th entry of that class's bit vector is 1. This results in a total of 7 datasets with sizes 3103, 2972, 3016, 3147, 3091, 2912, and 2972. The images belonging to classes not assigned a bit vector were discarded, leaving a total of 7071 remaining images.

For the splits used in the main experiment, we assigned bit vectors of length 16 with Hamming weight 8 (for MNIST), length 20 with Hamming weight 10 (for CIFAR-10), and length 24 with Hamming weight 12 (for CelebA). Bit vectors were sampled uniformly at random without replacement. This resulted in 16 splits for MNIST of sizes 4970, 4993, 5033, 5008, 5001, 4961, 5000, 5024, 4960, 5026, 4984, 4990, 5025, 5012, 4987, and 5026,
20 splits for CIFAR-10 of sizes 30053, 29960, 30027, 29998, 30006, 29992, 30056, 30030, 30009, 29962, 30196, 29780, 29932, 30059, 30138, 29925, 29967, 29870, 29985, and 30055, and 24 splits for CelebA of sizes 100841, 101578, 101254, 101473, 101092, 101143, 101271, 101229, 101596, 101270, 101455, 101358, 101224, 101402, 101294, 101054, 101788, 101207, 101539, 101026, 101339, 101614, 101131, and 101010.

\subsection{Model training}

We used the same model architecture as was used by \citet{ho2020denoising}. The PyTorch implementation of the model was sourced from \url{https://github.com/w86763777/pytorch-ddpm/blob/7abb4c0358db4f3f2b0a4609a9d726185b815e65/model.py}.

Training was performed as described in Algorithm 1 in the work of \citet{ho2020denoising}. Adam was used for descent with default pytorch parameters, with the exception of learning rate which was set to $2*10^{-4}$ for MNIST and CIFAR-10, and to $2*10^{-5}$ for CelebA. A minibatch size of 128 was used for training both MNIST and CIFAR-10 modelsm and a minibatch size of 32 was used for training CelebA models. MNIST training epochs consisted of minibatches sampled without replacement, and each model was trained for 501 epochs. CIFAR-10 training epochs consisted of 10000 minibatches sampled with replacement, and each model was trained for 25 epochs. CelebA ALSO training epochs consisted of 10000 minibatches sampled with replacement, and each model was trained for 13 epochs. Images used for training CIFAR-10 and CelebA models were also randomly flipped horizontally for data augmentation purposes.

\subsection{Experiments}
\label{appdx-training-experiments}

Sampling was performed as described in Algorithm 2  in the work of \citet{ho2020denoising}. At each denoising step, the predicted noise of all the models were averaged to produce a single output. Forward-mode Automatic Differentiation in the PyTorch Automatic Differentiation package was used to compute the Jacobian. We note that the API is in beta as of PyTorch version 2.0.0, the version used for our experiments.

3000 samples along with 7 counterfactuals each were generated for the MNIST ensemble of size 7. 382 samples along with Jacobians, 64 counterfactuals, and 16 single model outputs each were generated for the MNIST ensemble of size 16, and 128 samples along with Jacobians and 64 counterfactuals, and 20 single model outputs each each were generated for the CIFAR-10 ensemble.

LPIPS was calculated using the TorchMetrics package (\url{https://torchmetrics.readthedocs.io/en/stable/image/learned_perceptual_image_patch_similarity.html}), where a VGGNet was used. FID was calculated using the last layer of inception\_v3 included in the TorchVision package with weights from Inception\_V3\_Weights.DEFAULT. Images were rescaled from either 28-by-28 (for MNIST) or 32-by-32 (for CIFAR-10) to 299-by-299 before input with torchvision.transforms.Resize. The operation was applied to tensors with default parameters, except for antialias which we set to True.

\subsection{Resources}

The training and experiments were carried out over the course of 3 weeks on 8 GeForce GTX 1080 Ti (11GB) cores, 8 Titan RTX (24GB) cores, and 12 A100 (40GB) cores.

\bibliography{references}
\end{document}